%% file: main.tex
\documentclass[sigconf]{acmart}

\input{useful_package}

\AtBeginDocument{%
  }

\copyrightyear{2025}
\acmYear{2025}
\setcopyright{acmlicensed}\acmConference[MM '25]{Proceedings of the 33rd ACM International Conference on Multimedia}{October 27--31, 2025}{Dublin, Ireland}
\acmBooktitle{Proceedings of the 33rd ACM International Conference on Multimedia (MM '25), October 27--31, 2025, Dublin, Ireland} \acmDOI{10.1145/3746027.3755013}
\acmISBN{979-8-4007-2035-2/2025/10}

\makeatletter
\gdef\@copyrightpermission{
\begin{minipage}[c]{0.21\columnwidth}
\vspace{2pt}
  \href{https://creativecommons.org/licenses/by/4.0/}{%
    \includegraphics[width=0.90\textwidth]{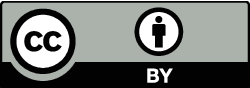}}
\vspace{2pt}
\end{minipage}\hfill
\begin{minipage}[c]{0.79\columnwidth}
  \href{https://creativecommons.org/licenses/by/4.0/}{%
    This work is licensed under a Creative Commons
    Attribution International 4.0 License.}
\end{minipage}
}
\makeatother

\begin{document}

\title{Positional Prompt Tuning for Efficient 3D Representation Learning}

\author{Shaochen Zhang}
\authornote{Co-first author.}
\orcid{0009-0006-5550-905X}
\affiliation{%
  \institution{Xi'an Jiaotong University}
  \city{Xi'an}
  \country{China}
}
\author{Zekun Qi}
\orcid{0009-0001-2554-5141}
\authornotemark[1]
\affiliation{%
  \institution{Tsinghua University}
  \city{Beijing}
  \country{China}
}
\author{Runpei Dong}
\orcid{0000-0002-1104-7897}
\affiliation{%
  \institution{University of Illinois Urbana-Champaign}
  \city{Champaign}
  \country{United States}
}
\author{Xiuxiu Bai}
\orcid{0000-0002-8102-1596}
\affiliation{%
  \institution{Xi'an Jiaotong University}
  \city{Xi'an}
  \country{China}
}
\author{Xing Wei}
\orcid{0000-0002-5025-3941}
\authornote{Corresponding author. \faIcon[regular]{envelope} weixing@xjtu.edu.cn}
\affiliation{%
  \institution{Xi'an Jiaotong University}
  \city{Xi'an}
  \country{China}
}


\begin{abstract}
We rethink the role of positional encoding in 3D representation learning and fine-tuning. We argue that using positional encoding in point Transformer-based methods serves to aggregate multi-scale features of point clouds. Additionally, we explore parameter-efficient fine-tuning (PEFT) through the lens of prompts and adapters, introducing a straightforward yet effective method called PPT for point cloud analysis. PPT incorporates increased patch tokens and trainable positional encoding while keeping most pre-trained model parameters frozen. Extensive experiments validate that PPT is both effective and efficient.
Our proposed method of PEFT tasks, namely PPT, with only 1.05M of parameters for training, gets state-of-the-art results in several mainstream datasets, such as 95.01\% accuracy in the ScanObjectNN OBJ\_BG dataset. Codes and weights will be released at 
\url{https://github.com/zsc000722/PPT}.
\end{abstract}




\maketitle

\input{chapters/introduction}
\input{chapters/related_work}
\input{chapters/method}
\input{chapters/experiments}

\input{chapters/discussions}
\input{chapters/future_work}
\input{chapters/conclusion}

\bibliographystyle{ACM-Reference-Format}
\bibliography{cite}

\input{chapters/appendix}

\end{document}

%% file: useful_package.tex
\usepackage[utf8]{inputenc} 
\usepackage[T1]{fontenc}    
\usepackage{url}            
\usepackage{amsfonts}       
\usepackage{nicefrac}       
\usepackage{microtype}      
\usepackage{xcolor}         
\usepackage{fontawesome5} 

\theoremstyle{plain}

\theoremstyle{definition}

\theoremstyle{remark}

\usepackage{microtype}
\usepackage{graphicx}
\usepackage{booktabs} 

\usepackage{cleveref}

 \makeatletter
\def\@fnsymbol#1{\ensuremath{\ifcase#1\or \dagger\or \ddagger\or
  \mathsection\or \mathparagraph\or \|\or **\or \dagger\dagger
  \or \ddagger\ddagger \else\@ctrerr\fi}}
\makeatother

\usepackage{multirow}
\usepackage{enumerate}
\usepackage{colortbl}
\usepackage{enumitem}
\usepackage{wrapfig}
\usepackage{bbding}
\usepackage{pifont}
\usepackage{wasysym}
\usepackage{textcomp}

\usepackage{color}
\def\eg{{\it{e.g.}}}

\def\etc{{\it{etc}}}

\def\vpp{{\scshape VPP}}
\def\recon{{\scshape ReCon}}
\def\act{{\scshape ACT}}
\def\ppt{{\scshape PPT}}

\definecolor{pretty-blue}{RGB}{0, 113, 188}
\definecolor{linecolor}{rgb}{0.82, 0.94, 0.75}
\definecolor{trainable_red}{RGB}{220, 50, 80}
\definecolor{table_row}{RGB}{230, 244, 236}
\definecolor{trainable_blue}{RGB}{10, 120, 220}

\usepackage{caption}
\usepackage{float}

\usepackage{tabularx}
\usepackage{adjustbox}

\usepackage{footnote}
\usepackage{url}
\usepackage[flushleft]{threeparttable}
\usepackage{enumitem}
\usepackage{algorithm}
\usepackage{algorithmic}
\usepackage{listings}
\usepackage{tabularx}
\usepackage{pifont} 
\usepackage{bm}
\usepackage{tablefootnote}
\usepackage{tcolorbox}

\usepackage{bibentry}
\usepackage{booktabs}

%% file: chapters/introduction.tex
\vspace{-0.3em}
\section{Introduction}
With the increasing popularity of scanning devices such as RGBD cameras and LiDAR, we are witnessing the rise of a new modality of data: 3D point clouds. It has a wide range of applications for many tasks, such as autonomous driving and robot grasping, \etc. Because of the huge potential for these downstream tasks, technological updates in 3D point cloud representation learning are iterating very fast, from PointNet~\cite{PointNet}, PointNet++~\cite{PointNet++} and then all the way down to the Transformer based methods~\cite{PointMAE, PointBERT, PointM2AE22, PointTransformer,recon23,han2024mamba3d} which now occupying the dominant position in this field. Within all these Transformer-based 3D representation learning methods, Position Encoding plays an important role in offering location specificity for the Transformer architecture, which can not distinguish tokens from different positions. Different from other domains like vision or language, the position encoding in the point cloud domain is usually not designed elaborately, like the rotary position embedding\citep{su2024roformer}, which is now widely adopted in the NLP domain to implement relative position embedding in an absolute manner. Instead, a lightweight MLP with cluster centers as input serves as the position encoding module. Noticing this circumstance, we then wonder \textbf{why a simple MLP as position encoding can efficiently facilitate the performance in 3D representation learning?}
\input{figs/analysis}

\input{figs/compare}
We hold the opinion that the position encoding MLP, taking center points as input, together with the patch encoder, which deals with the neighbor points around the cluster centers, compose a multi-scale information extractor. Then, the multi-scale patch embeddings of points are fed into the Transformer for subsequent feature extraction. The Max or Mean Pooling operation doesn't break the disordered and continuous property. Thus, this multi-scale method, which focuses on both local and global features, gains excellent performance in 3D representation learning. As shown in ~\cref{fig:poster}(b), the average attention distance of position encoding patterns is relatively centralized, while the local patch patterns hold a dispersed attention distribution, indicating the diverse concern levels of position encoding and patch tokens. In the meantime, the position encoding itself only needs a few parameters ($\sim$0.2$\%$), so it is suitable for promoting the performance of PEFT tasks. 

PEFT methods usually freeze most parameters of pre-trained models, only a few selected parameters and some other ones inserted are trainable during tuning, therefore significantly releasing the demand for computational resources of full fin-tuning. In the meantime, by only partly saving the trainable parameters, the pressure of storage space is also released. For language or vision models, there are Adapter tuning~\cite{adaptertuning,10223170,li2024adapter,zhao2024dynamic}, Prompt tuning~\cite{prompttuning, VPT22}, and LoRA~\cite{LoRA}, which introduce extra trainable layers or prompts into pre-trained models and get results comparable or even better than full fine-tuning. So it is natural to implement the same approach on 3D point cloud models. We attach importance to the positional embedding of point cloud representation as well as fine-tuning, unfreezing the position encoding part in the fine-tuning process, and thus get better results.

In the meantime, we draw our attention to plenty of different prompt-based PEFT methods in the point cloud analysis domain~\cite{DAPT,IDPT}, which share a similar paradigm. Each one of these methods inserts prompts into the encoder inputs. Instead of static prompt tokens that are selected manually, these prompts are dynamically generated by elaborate structures like the DGCNN layer in IDPT or linear layers followed by the activation function in DAPT. Both of them use an adapter-like way to generate extra prompts from the Transformer layer outputs, which brings them satisfactory results. Nevertheless, considering the prompts in the NLP domain, they are linguistically meaningful, like a series of adjectives depicting the fine-grained characteristics of the object or a specific description for the exact question. The aforementioned adapter-generated prompts are more likely to aggregate the information, thus, there is nothing new for the model to get from the prompts. 

Nevertheless, our PPT adopts encoded sampling centers, together with trainable positional embedding as extra prompt tokens, as shown in \cref{fig:compare}. There is a fundamental distinction between this form of prompt and the previously discussed aggregate-based prompts. These prompts, obtained through sampling and clustering methods, inherently carry physical meaning in real space: specifically, the three-dimensional positional information of points. By inputting such prompts into subsequent networks and dynamically adjusting them through adapters between the Transformer layers, we can better generalize the information from pre-trained encoders to downstream tasks during training.
Our entire network architecture is remarkably straightforward and can be seamlessly integrated into transferring tasks of any Transformer-based point cloud pre-trained model. This indicates that our structure is highly reusable and holds significant potential for improvement.

Considering these perspectives, we propose our Positional Prompt Tuning, namely PPT. With only a few trainable parameters, we emphasize the importance of positional embedding layers and set them trainable, so as the initial encoder, which transfers point clouds into patch tokens. We also insert simple trainable adapter layers between each layer of the Transformer encoder to adjust the weighted features dynamically.

Our main contributions can be summarized as follows:
\begin{itemize}[leftmargin=13pt]
    \item We emphasize the significance of positional embedding in 3D point cloud representation learning, and conduct extensive experiments to validate its effectiveness.
    \item We revisit the Parameter-efficient Prompt Tuning problem in a Prompt and Adapter manner and propose a quite simple method, Positional Prompt Tuning (PPT), combining the above two points. 
    \item With only \textbf{5$\%$} of the trainable parameters compared with the full fine-tuning, our PPT significantly reduces the demand for storage space during fine-tuning. Meanwhile, extensive experiments have also shown that our PPT substantially reduces the time of fine-tuning while achieving on par or even higher performance, \eg, 1.26$\%$ accuracy increasing on ReCon under ModelNet 1k point, and 4.82$\%$ on Point-MAE under ScanObjectNN dataset.
\end{itemize}

%% file: figs/analysis.tex
\begin{figure}[t] 
\centering
\vspace{12pt}
\includegraphics[width=1\linewidth]{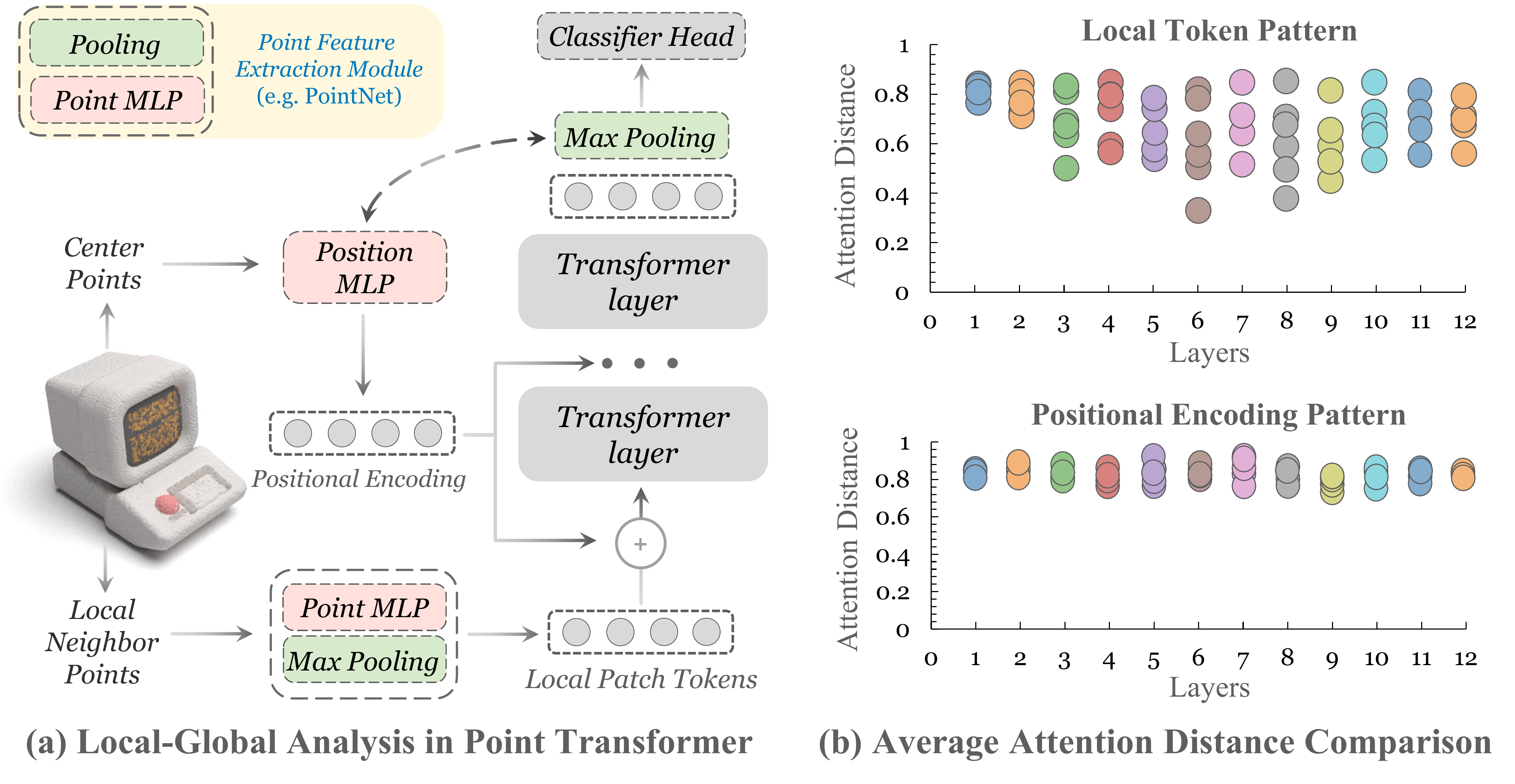}
\vspace{-20pt}
\caption{\textbf{The Role of Position Encoding in Transformers.}}
\vspace{-12pt}
\label{fig:poster}
\end{figure}

%% file: figs/compare.tex
\begin{figure*}[t] 
    \centering  
    \includegraphics[width=1.0\linewidth]{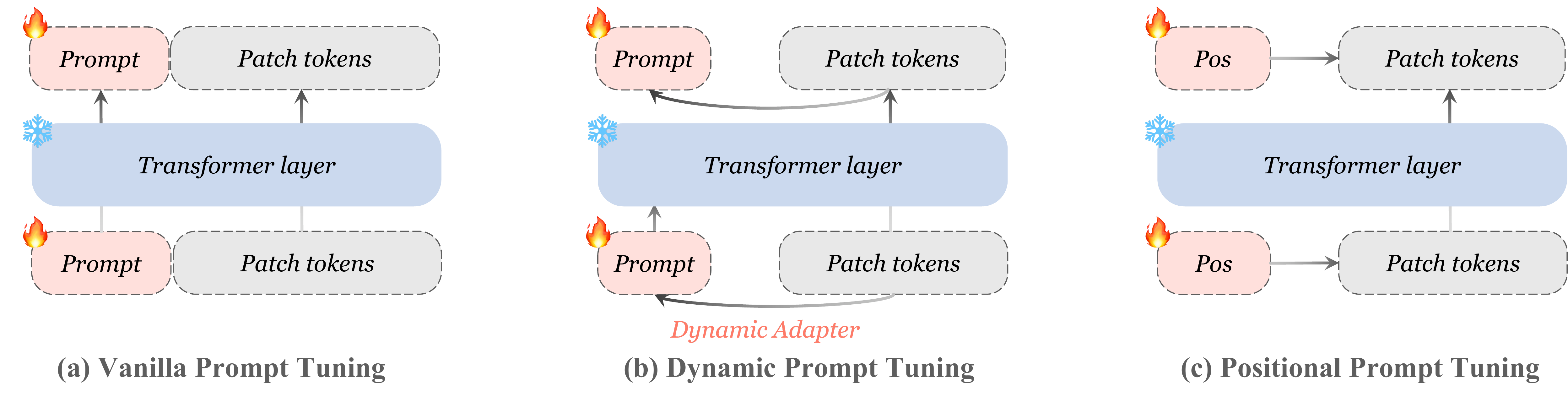}
    \vspace{-18pt}
    \caption{Comparison of three different prompt methods. The vanilla prompt tuning methods in (a) straightly add trainable parameters as prompts, while the dynamic prompt tuning methods in (b) adopt trainable extra dynamic adapters to generate prompts. Our PPT in (c) adopts trainable positional embedding for prompts.}
    \label{fig:compare}
    \vspace{2pt}
\end{figure*}

%% file: chapters/related_work.tex
\input{figs/pipeline}
\section{Related Works}
\subsection{3D Representation Learning}
3D Representation Learning includes point-based~\cite{PointNet, PointNet++}, voxel-based~\cite{voxelnet15}, and multiview-based methods~\cite{mvcnn15,MVTN}, \etc. Due to the sparse but geometry-informative representation, point-based methods~\cite{PointNext, PointTrans21} have become mainstream approaches in object classification~\cite{ModelNet15,ScanObjectNN19}. Voxel-based CNN methods~\cite{SyncSpecCNN17,voxelrcnn21} provide dense representation and translation invariance, achieving outstanding performance in object detection~\cite{ScanNet17} and segmentation~\cite{ShapeNetPart16,S3DIS16}. Furthermore, due to the vigorous development of attention mechanisms~\cite{AttentionIsAllYouNeed}, 3D Transformers~\cite{PointTrans21,groupfree21,voxeltransformer21} have also brought about effective representations for downstream tasks. Recently, 3D self-supervised representation learning has been widely studied\cite{liang2024pointmamba,chen2024sugar,tang2024any2point,zha2024towards,zhang2024pcp,he2024pointrwkv,ma2024shapesplat,shen2024diffclip,luo2024learning,song2025mv,sun2024parameter,wang2024multi}. PointContrast~\cite{PointContrast20} leverages contrastive learning across different views to acquire discriminative 3D scene representations. Point-BERT~\cite{PointBERT} and Point-MAE~\cite{PointMAE} first introduce masked modeling pretraining into 3D. \act~\cite{ACT23} pioneers cross-modal geometry understanding via 2D/language foundation models. \recon~\cite{recon23, shapellm24} further proposes to unify generative and contrastive learning.
Facilitated by foundation vision-language models like CLIP~\cite{CLIP}, another line of works are proposed towards open-world 3D representation learning~\cite{ULIP22,OpenScene23,CLIPFO3D23,LAS3D23,PLA23,PointGCC23,vpp23,xia2025towards}.

\subsection{Parameter-Efficient Fine-Tuning}
The pre-training and fine-tuning paradigm has become a mainstream approach in large models to maximize their semantic understanding ability for solving downstream tasks in multiple subdivisions. However, the full fine-tuning paradigm used in most cases consumes considerable storage and may also lead to catastrophic forgetting, degrading model performance. To solve this problem, researchers in the field of NLP and 2D have proposed many effective methods as PEFT (Parameter-Efficient Fine-tuning). 
Prompt Tuning~\cite{prompttuning} and Prefix tuning~\cite{prefixtuning} introduce extra inputs or vectors as the only trainable parts, while Adapter tuning~\cite{adaptertuning} inserts additional modules between layers as the trainable part. LoRA~\cite{LoRA} adopts low-rank approximation matrices in a parallel manner to simulate the full fine-tuning process. VPT~\cite{VPT22} then introduces prompt tuning modules into pre-trained image models. These methods have achieved substantial achievements in their own specialty. So far, in the 3D domain, several methods discussing the PEFT on point cloud pre-train models demonstrate their effectiveness~\cite{IDPT,DAPT,PointPEFT,liang2024parameter,han2025most,fei2024fine}. IDPT~\cite{IDPT} generates instance-aware dynamic prompts instead of static prompts with a DGCNN module. DAPT~\cite{DAPT} then caps every layer with a light-weight TFTS layer and uses dynamic adapters to generate prompts, internally inserting them in the inputs of encoders. Point-PEFT~\cite{PointPEFT} constructs a point-prior bank to store feature templates of training data and uses a parameter-free attention mechanism for feature aggregation to generate prompts into the first $\mathcal{L}$ encoder layers.

%% file: figs/pipeline.tex
\begin{figure*}[t] 
    \centering  
    \includegraphics[width=1\linewidth]{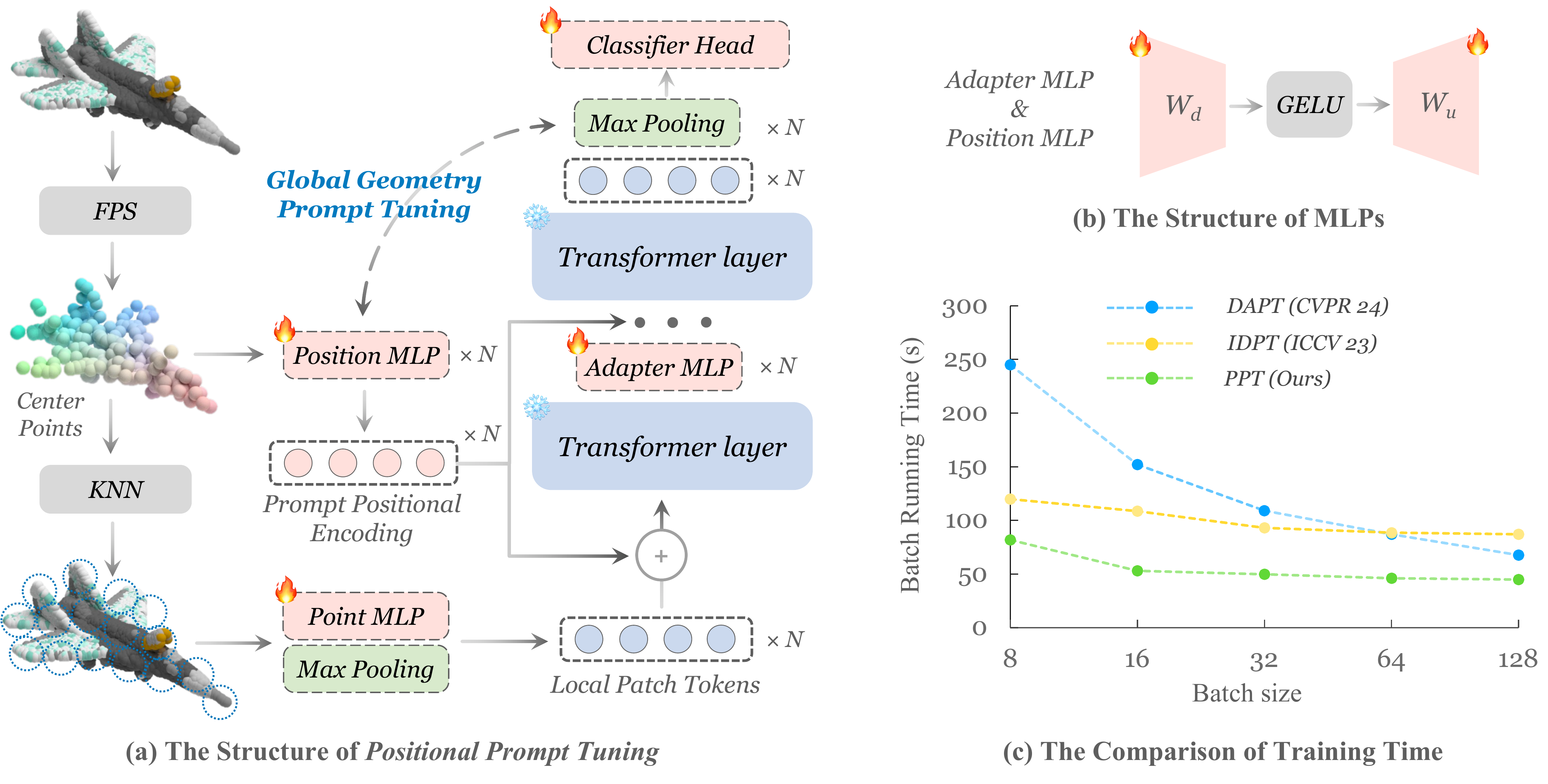}
    \vspace{-15pt}
    \caption{\textbf{Training overview of \ppt.} (a) The structure of \textit{Positional Prompt Tuning} (\ppt). (b) We use a simple FFN structure as a position MLP and an adapter MLP. (c) The training time comparison between IDPT~\cite{IDPT}, DAPT~\cite{DAPT} and our \ppt.}
    \label{fig:training}
    \vspace{2pt}
\end{figure*}

%% file: chapters/method.tex
\section{\ppt: Positional Prompt Tuning}
The existing methods of PEFT mostly summarize and transfer some existing ways from other fields to the current domain and improve them to get better results. We attach importance to positional embedding in 3D missions and emphasize the role of position information in 3D tasks and then revisit the PEFT problem from the perspective of prompts and adapters. With several simple modifications of the network, we propose our \ppt, Positional Prompt Tuning method, which comprehensively considers both aspects above and achieves excellent results. In this section, we discuss the specifics of our \ppt~as follows.

\subsection{Why does positional prompt matter?}
\label{method_pos}

Positional encoding has always been a vital part of all Transformer-based structures, for their input form naturally lacks positional information. Unlike the comprehensive research in NLP~\cite{su2024roformer} and 2D vision~\cite{wu2021rethinking} fields, the positional encoding pattern is not as well-designed for the point cloud domain. Mostly in the 3D domain, the positional information is aggregated by passing patch centers into a simple MLP and added to the input patch tokens~\cite{PointMAE, PointBERT}. Since the input of the position encoding layer is the cluster centers of the target point cloud, this highly semantic-rich input form can achieve excellent results with barely a simple MLP layer, even better than some commonly used position encoding forms in NLP, such as sinusoidal position encoding or RoPE. In fact, after re-examining the location information in 3D coped with MLP, we find that this form of position encoding is actually a multi-scale feature combined with the information of patch tokens, or to say, is a \textit{little PointNet}. Due to the disorderly property of point clouds, the current mainstream point cloud feature extraction network, such as PointNet~\cite{PointNet}, generally adopts the following structure when a point cloud set $X\in \mathbb{R}^{N\times 3}$ is given:
\begin{equation}
    F_{pc}(X) = L_{M}(\gamma (X))
\end{equation}
Where $\gamma$ usually represents an MLP or a convolutional layer as point point-level feature extractor and $L_M$ denotes a max or mean pooling layer. Due to the disordered and continuous characteristics of point clouds, a sequence-independent feature aggregation layer like max or mean pooling is usually adopted, as adding and maximizing don't break this property. In the point Transformer-based methods, a group of centers' coordinates are embedded and then added to the patch token inputs as positional information. With $g$ center points $X_c = x_{c_1}, x_{c_2}, ..., x_{c_g}, x_{c_i}\in\mathbb{R}^{3}$ selected by methods like FPS and their neighbor points $X_{n} = X_{c_1}, X_{c_2}, ..., X_{c_g}, X_{c_i}\in\mathbb{R}^{g\times 3}$ are given, their basic structure are as follows:
\begin{equation}
    E_{pt} = \gamma_{n}(X_{c_1}, X_{c_2}, ..., X_{c_g}), 
\end{equation}
\begin{equation}
    E_{pos} = \gamma_{c}(x_{c_1}, x_{c_2}, ..., x_{c_g})
\end{equation}
\begin{equation}
    f_{x} = L_M(\Psi(E_{pt}+E_{pos}))
\end{equation}
where $\Psi$ denotes the Transformer Encoder as the feature extractor, and $\gamma_c, \gamma_n$ denotes MLP encoder layer. We separately investigate the two parts of the input of the encoder, which are originally added together. 
\begin{equation}
    f_c = L_{M}(\Psi(E_{pos})))
\end{equation}
\begin{equation}
    f_n = L_{M}(\Psi(E_{pt}))
\end{equation}
The $f_c$ branch accepts embedded center coordinates as input, which are more global messages, and together with the $f_n$ branch, it takes in more local and detailed messages from neighbor points. Integrating these two different branches together, the $f_x$ structure is capable of mixing information of different scales. Thus, the final output of the encoder, then copied with a max or mean pooling layer, is actually a multi-scale feature extractor of point cloud, holding the same effect as both utilize $f_c$ and $f_n$. This multi-scale feature, rich in semantic information, enables the model to understand local features and at the same time to accept global location information.

\input{tabs/alg}
\input{tabs/cls}

\subsection{Better way of prompt and adapter tuning?}
\label{method_moe}
 With point cloud as the only modality of input, it is difficult for 3D parameter-efficient fine-tuning tasks to get prompts that are rich in linguistic significance, like a well-designed prefix prompt in natural language, which can remarkably boost the performance of generative tasks in text modality~\cite{prompttuning}. VPT~\cite{VPT22} introduces a set of trainable embeddings into the input of the Transformer to solve this problem in the vision domain, and both DAPT~\cite{DAPT} and IDPT~\cite{IDPT} adopt a similar manner. Simultaneously considering Adapter and Prompt, they both generate extra prompt tokens from an adapter-structure network and add these prompt tokens into the  Transformer input.
 
 By doing so, they acquire dynamic prompts rather than static ones in VPT. As the static prompts always need a process of manual selection, the generated dynamic prompts are a compromise of this process. Nevertheless, these kinds of prompts are actually an abstraction of the existing feature embedding. Thus the input of Transformer layers with extra prompt inserted $d_{in}$ and the Transformer output $d_{out}$can be formulated as:
 \begin{equation}
     d_{in} = [\textcolor{trainable_red}{CLS}, \textcolor{trainable_red}{\varPsi_a}(P_t), P_t]
 \end{equation}
 \begin{equation}
     d_{out} = \textcolor{trainable_blue}{F}(\textcolor{trainable_blue}{attn}(d_{in}))
 \end{equation}
\noindent Where \textcolor{trainable_red}{$CLS$} means class token, \textcolor{trainable_red}{$\varPsi_a$}$(P_t)$ represents the extra prompts abstracted from patch embedding $P_t$. \textcolor{trainable_blue}{attn} and \textcolor{trainable_blue}{F} represent the FFN and attention layer inside the Transformer blocks. The $\textcolor{trainable_red}{\bullet}$ and $\textcolor{trainable_blue}{\bullet}$ represent \textcolor{trainable_red}{trainable} and \textcolor{trainable_blue}{frozen} parameters, respectively. Thus, the information of prompts is highly homogeneous, which is an aggregation of existing tokens, rather than the newly added language prompts in the NLP domain. We then think about whether there are some other forms of prompts that can both be dynamically adjusted and, in the meantime, be of physical significance. And it comes to our mind that the patch tokens themselves, which are down-sampled and clustered, can be used as extra tokens for prompting. Different from prompts that are yielded from patch token embedding, we unfreeze a part of the patch encoder to introduce variety into patch token embedding. In that case, with this form of prompts, the input of Transformer layers $f_{in}$ can be formulated as:
\begin{equation}
    f_{in} = [\textcolor{trainable_red}{CLS}, \textcolor{trainable_red}{\varphi_{a1}}(P_{t_1}), \textcolor{trainable_red}{\varphi_{a2}}(P_{t_2})]
\end{equation}
 Here $\textcolor{trainable_red}{\varphi_{a1}}$ and $\textcolor{trainable_red}{\varphi_{a2}}$ mean different adapter layers for dynamic adjustment. Thus the output of the Transformer $f_{out}$ is:
\begin{equation}
    f_{out} = \textcolor{trainable_blue}{F}(\textcolor{trainable_red}{\xi_{a}}(\textcolor{trainable_blue}{attn}(f_{in})))
\end{equation}
\noindent\textcolor{trainable_red}{$\xi$} represents the added trainable adapter for more thorough feature aggregation and adjustment. In this way, both the extra sampled patch token prompts and the original patch token embedding are altered for information integration. In this manner of prompts and adapters, combined with the aforementioned positional encoding, we carried out extensive experiments, and our method achieved excellent results in all of these experiments, which will be discussed in subsequent chapters. 

\input{tabs/fewshot}

\subsection{Pipeline of \ppt}
The pipeline structure of our \ppt\ is shown in \cref{fig:training}(a). The patch encoder of the pre-trained model is already empowered to abstract a high concentration of semantic information from the input point cloud, so we unfreeze part of the patch encoder to provide variety for patch tokens and prompt tokens. Briefly, we resample the input point clouds as prompts and concatenate them with the input patch tokens before sending them into the Transformer, and trainable positional embedding layers are concatenated together as the positional information for both the prompts and the patch tokens. Before inputting it into each Transformer layer, a patch adapter layer and a prompt adapter layer are adopted as a dynamic adjustment module. There are also adapter layers inside every Transformer block, specifically between the attention layer and the FFN. As the pre-trained parts of the Transformer layers are untrainable during the tuning process, an adapter inside the block effectively integrates information and dynamically adjusts it to improve performance.

\subsection{Algorithm Pseudo Coding}
In \cref{alg:DA}, we report the pseudo-code of our algorithm. As we can see from the fairly simple code, the overall implementation process is concise but very efficient.

%% file: tabs/alg.tex
\begin{algorithm}[H]
\small
\caption{\small Positional Prompt Tuning.}
\label{alg:DA}
\definecolor{codeblue}{rgb}{0.25,0.5,0.5}
\definecolor{codekw}{rgb}{0.85, 0.18, 0.50}
\lstset{
  backgroundcolor=\color{white},
  basicstyle=\fontsize{8.6pt}{8.6pt}\ttfamily\selectfont,
  columns=fullflexible,
  breaklines=true,
  captionpos=b,
  commentstyle=\fontsize{10pt}{10pt}\color{codeblue},
  keywordstyle=\fontsize{10pt}{10pt}\color{codekw},
  xleftmargin=4pt,
  escapechar={|},
  numbers=none,
}
\begin{minipage}{\textwidth}
\begin{lstlisting}[language=python]
# An example of Positional Prompt Tuning
import torch.nn as nn
import torch.nn.functional as F

class PositionalPromptTuning:
    def __init__():
        norm = LayerNorm()
        group = Group(num, size)
        encoder = PatchEncoder(grad=True)
        cls_token = Parameter()
        cls_pos = Parameter()
        pos_embed = MLP(grad=True)
        block = TransformerEncoder()
        finetune_head = MLPHead(grad=True)
    
    def forward(points):
        neighbor, center = group(points)
        patch = encoder(neighbor)
        pos = pos_embed(center)
        extra_pos = extra_pos_embed(extra_ctr)
        
        # Positional Prompt Tuning
        x = concat(cls_token, patch, patch)
        pos = concat(cls_pos, pos, extra_pos)
        
        x = block(x, pos)
        x = norm(x)
        out = finetune_head(x)
        
        return out

\end{lstlisting}
\end{minipage}
\end{algorithm}
\vspace{-5pt}

%% file: tabs/cls.tex
\begin{table*}[t!]
\setlength\tabcolsep{4.2pt}
\caption{
\textbf{Classification on three variants of the ScanObjectNN~\cite{ScanObjectNN19} and the ModelNet40~\cite{ModelNet15}}, including the number of trainable parameters and overall accuracy (OA). All methods utilize the default data argumentation as the baseline. \textcolor{red}{$^*$} denotes reproduced results. We report the highest and average values by running eight iterations, referred to (-/-). For a fair comparison, we report all the results \textbf{without the voting strategy}~\cite{RSCNN}.}
\label{tab:cls}
\begin{center}
\resizebox{1.0\linewidth}{!}{
\begin{tabular}{lccccccc}
\toprule[0.95pt]
\multirow{2}{*}[-0.5ex]{Method}  &\multirow{2.3}{*}{Reference} & \multirow{2}{*}[-0.5ex]{ Params. (M)} & \multicolumn{3}{c}{ScanObjectNN} & \multicolumn{2}{c}{ModelNet40}\\
\cmidrule(lr){4-6}\cmidrule(lr){7-8} & & & OBJ\_BG & OBJ\_ONLY & PB\_T50\_RS& 1k P & 8k P\\
\midrule[0.6pt]
\multicolumn{8}{c}{\textit{Supervised Learning Only}}\\
\midrule[0.6pt]
PointNet~\cite{PointNet} & CVPR 17 & 3.5 & 73.3 & 79.2 & 68.0 & 89.2 & 90.8\\
PointNet++~\cite{PointNet++} & NeurIPS 17 & 1.5 & 82.3 & 84.3 & 77.9 & 90.7 & 91.9\\
DGCNN~\cite{DGCNN} & TOG 19 & 1.8 & 82.8 & 86.2 & 78.1 & 92.9 & -\\
PCT~\cite{PCT} & ICCV 21 & 2.88 & - & - & - & 93.2 & -\\
PointMLP~\cite{PointMLP} & ICLR 22 & 12.6 & - & - & 85.4$\pm$0.3 & 94.5 & -\\
PointNeXt~\cite{PointNext} & NeurIPS 22 & 1.4 & - & - & 87.7$\pm$0.4 & 94.0 & -\\
\midrule[0.6pt]
\multicolumn{8}{c}{\textit{with Self-Supervised Representation Learning} ({\scshape Full})}\\
\midrule[0.6pt]
OcCo~\cite{occo} & ICCV 21 & 22.1 & 84.85 & 85.54 & 78.79 & 1k & - / 92.1 \\
Point-BERT~\cite{PointBERT} & CVPR 22 & 22.1 & 87.43 & 88.12 & 83.07 & 93.2 & 93.8\\
Point-MAE~\cite{PointMAE} & ECCV 22 & 22.1 & 90.02 & 88.29 & 85.18 & 93.8 & 94.0\\
Point-M2AE~\cite{PointM2AE22} & NeurIPS 22 & 15.3 & 91.22 & 88.81 & 86.43 & 94.0 & -\\
\act~\cite{ACT23} & ICLR 23 & 22.1 & 93.29 & 91.91 & 88.21 & 93.7 & 94.0\\
\vpp~\cite{vpp23} & NeurIPS 23 & 22.1 & 93.11 & 91.91 & 89.28 & 94.1 & 94.3\\
I2P-MAE~\cite{I2PMAE23} & CVPR 23 & 15.3 & 94.15 & 91.57 & 90.11 & 94.1 & -\\
\recon~\cite{recon23} & ICML 23 & 44.3 & 95.18 & 93.29 & 90.63 & 94.1 & 94.3\\
\midrule[0.6pt]
\multicolumn{8}{c}{\textit{with Parameter-Efficient Supervised Fine-tuning}}\\
\midrule[0.6pt]
Point-MAE~\cite{PointMAE} (Baseline) & ECCV 22 & 22.1 & 90.02 & 88.29 & 85.18 & 93.8 & 94.0\\
+ IDPT\textcolor{red}{$^*$}~\cite{IDPT} & ICCV 23 & 1.7 & 
92.94/\textcolor{gray}{92.38} & 
92.60/\textcolor{gray}{91.33} & 
88.34/\textcolor{gray}{88.09} & 
93.64/\textcolor{gray}{93.02} & 
93.88/\textcolor{gray}{93.67}\\
+ DAPT\textcolor{red}{$^*$}~\cite{DAPT} & CVPR 24 & 1.1 & 
92.43/\textcolor{gray}{91.61} & 
91.91/\textcolor{gray}{91.31} & 
88.27/\textcolor{gray}{87.68} & 
92.99/\textcolor{gray}{92.83} & 
93.27/\textcolor{gray}{93.05}\\
\rowcolor{table_row}+ \ppt~(Ours) & ACMMM 25 & 1.1 & 
\textbf{93.63}/\textcolor{gray}{\textbf{92.99}} & 
\textbf{92.60}/\textcolor{gray}{\textbf{92.43}} & 
\textbf{89.00}/\textcolor{gray}{\textbf{88.41}} & 
\textbf{93.68}/\textcolor{gray}{\textbf{93.18}} & 
\textbf{93.88}/\textcolor{gray}{93.51}\\
\midrule[0.6pt]
\recon~\cite{recon23} (Baseline) & ICML 23 & 22.1 & 94.32 & 92.77 & 90.01 &  92.5 & 93.0\\
+ IDPT\textcolor{red}{$^*$}~\cite{IDPT} & ICCV 23 & 1.7 & 
93.46/\textcolor{gray}{92.88} & 
91.74/\textcolor{gray}{91.37} & 
88.13/\textcolor{gray}{87.76} & 
93.64/\textcolor{gray}{93.28} &
93.64/\textcolor{gray}{93.52}\\
+ DAPT\textcolor{red}{$^*$}~\cite{DAPT} & CVPR 24 & 1.1 & 93.63/\textcolor{gray}{93.12} & 
92.43/\textcolor{gray}{91.63} & 
89.31/\textcolor{gray}{88.77} & 
93.27/\textcolor{gray}{92.94} & 
93.07/\textcolor{gray}{92.79}\\
\rowcolor{table_row}+ \ppt~(Ours) & ACMMM 25 & 1.1 & \textbf{95.01}/\textcolor{gray}{\textbf{94.09}} & \textbf{93.28}/\textcolor{gray}{\textbf{93.23}} & \textbf{89.52}/\textcolor{gray}{\textbf{88.97}} &
\textbf{93.76}/\textcolor{gray}{\textbf{93.41}}& 
\textbf{93.84}/\textcolor{gray}{\textbf{93.66}}\\
\bottomrule[0.95pt]
\end{tabular}
}
\end{center}
\vspace{5pt}
\end{table*}

%% file: tabs/fewshot.tex
\begin{table*}[t!]
\centering
\caption{\textbf{Few-shot Learning on ModelNet40}. Overall accuracy (\%) is reported. \textcolor{red}{$^*$} denotes reproduced results. We use the same environment and seed to ensure reproducibility.}
\label{tab:few-shot}
\setlength\tabcolsep{10pt}
\begin{tabular}{lccccc}
\toprule[0.95pt]
\multirow{2}{*}[-0.5ex]{Method}& \multirow{2.3}{*}{Reference}& \multicolumn{2}{c}{5-way} & \multicolumn{2}{c}{10-way}\\
\cmidrule(lr){3-4}\cmidrule(lr){5-6} & & 10-shot & 20-shot & 10-shot & 20-shot\\
\midrule[0.6pt]
DGCNN~\cite{DGCNN} & TOG 19 &31.6 $\pm$ 2.8 &  40.8 $\pm$ 4.6&  19.9 $\pm$  2.1& 16.9 $\pm$ 1.5\\
OcCo~\cite{occo} & ICCV 21 &90.6 $\pm$ 2.8 & 92.5 $\pm$ 1.9 &82.9 $\pm$ 1.3 &86.5 $\pm$ 2.2\\
\midrule[0.6pt]
\multicolumn{6}{c}{\textit{with Self-Supervised Representation Learning} ({\scshape Full})}\\
\midrule[0.6pt]
OcCo~\cite{occo}& ICCV 21 & 94.0 $\pm$ 3.6& 95.9 $\pm$ 2.3 & 89.4 $\pm$ 5.1 & 92.4 $\pm$ 4.6 \\
Point-BERT~\cite{PointBERT} & CVPR 22 & 94.6 $\pm$ 3.1 & 96.3 $\pm$ 2.7 &  91.0 $\pm$ 5.4 & 92.7 $\pm$ 5.1\\
Point-MAE~\cite{PointMAE} & ECCV 22 & 96.3 $\pm$ 2.5&97.8 $\pm$ 1.8 & 92.6 $\pm$ 4.1 & 95.0 $\pm$ 3.0\\
Point-M2AE~\cite{PointM2AE22} & NeurIPS 22 & 96.8 $\pm$ 1.8&98.3 $\pm$ 1.4 & 92.3 $\pm$ 4.5 & 95.0 $\pm$ 3.0\\
\act~\cite{ACT23} & ICLR 23 & 96.8 $\pm$ 2.3 & 98.0 $\pm$ 1.4 & 93.3 $\pm$ 4.0 & 95.6 $\pm$ 2.8\\
\vpp~\cite{vpp23} & NeurIPS 23 & 96.9 $\pm$ 1.9 & 98.3 $\pm$ 1.5 & 93.0 $\pm$ 4.0 & 95.4 $\pm$ 3.1 \\
I2P-MAE~\cite{I2PMAE23} & CVPR 23 & 97.0 $\pm$ 1.8&98.3 $\pm$ 1.3 & 92.6 $\pm$ 5.0 & 95.5 $\pm$ 3.0\\
\recon~\cite{recon23}(baseline) & ICML 23 & 97.3 $\pm$ 1.9 & 98.9 $\pm$ 1.2 & 93.3 $\pm$ 3.9 & 95.8 $\pm$ 3.0 \\
\midrule[0.6pt]
\multicolumn{6}{c}{\textit{with Parameter-Efficient Supervised Fine-tuning}}\\
\midrule[0.6pt]
 \textbf{\recon\,w/ IDPT}\textcolor{red}{$^*$}~\cite{IDPT} & ICCV 23 & 96.9$\pm$2.4 & 98.3$\pm$0.7 & \textbf{92.8}$\pm$4.0 & 95.5$\pm$3.2\\
 \textbf{\recon\,w/ DAPT}\textcolor{red}{$^*$}~\cite{DAPT} & CVPR 24 & 95.6$\pm$2.8 & 97.7$\pm$1.6 & 91.9$\pm$4.1 & 94.6$\pm$3.5\\
 \rowcolor{table_row}\textbf{\recon\,w/ \ppt~(Ours)} & ACMMM 25 & \textbf{97.0}$\pm$ 2.7 & \textbf{98.7}$\pm$1.6 & 92.2$\pm$5.0 & \textbf{95.6}$\pm$2.9\\
\bottomrule[0.95pt]
\end{tabular}
\vspace{5pt}
\end{table*}

%% file: chapters/experiments.tex
\section{Experiments}

\subsection{3D Real-World Object Recognition}
ScanObjectNN~\cite{ScanObjectNN19} is one of the most challenging 3D datasets, which covers $\sim$15K real-world objects from 15 categories.
For a fair comparison, we report the results with and without the voting strategy~\cite{RSCNN} separately. 
The results of the object classification tasks of ScanObjectNN (OBJ\_BG, OBJ\_ONLY, PB\_T50\_RS) are shown in \cref{tab:cls}. For each variant, we conducted experiments 8 times with different random seeds and reported the best and the average scores, and it can be observed that: 
(i)Holistically, with only 1.05M, 4.7\% parameters of {\scshape Full} fine-tuning paradigm, the performance of our block is improved by +0.7\% and +11.3\% compared with that of standard Transformer under {\scshape Full} tuning protocol on {\scshape ReCon} and Point-MAE, respectively. 
(ii) Compared with the current best methods, IDPT and DAPT, our PPT also gets better results than both of them. Specifically, we achieve 2.4\% and 4.5\% performance improvement compared with DAPT and IDPT, respectively. That is, with PPT on both ReCon and Point-MAE achieve state-of-the-art performance.

\input{tabs/segmentation}
\subsection{3D Synthetic Object Recognition}
ModelNet~\cite{ModelNet15} is one of the most classical datasets for synthetic 3D object recognition. It contains $\sim$12K meshed 3D CAD objects of 40 (ModelNet40) or 10 (ModelNet10) categories.
We conducted the evaluation on the ModelNet40 dataset, including fine-tuning and few-shot learning. 
During training and testing, we use \textit{Scale\&Translate} as data augmentation in training following~\cite{PointNet, PointNet++}. 
The results are shown in \cref{tab:cls} and \cref{tab:few-shot}, respectively. We report two different settings with 1k points and 8k points, respectively. It can be observed that
(i) With only a few trainable parameters, compared with {\scshape Full} tuning protocol, {\scshape ReCon} gains 2.1\% performance improvement with our PPT. Point-MAE also gets results on par with the {\scshape Full} tuning with less than 5\% of the parameters trainable. 
(ii) Compared with the current best methods, IDPT and DAPT, with our PPT, we achieve 1.26\% and 0.32\% improvements higher than these two, respectively. Point-MAE with PPT achieves the best result on ModelNet40 with 93.88\% accuracy. And {\scshape ReCon} with PPT gains +1.3\% and +0.8\% performance improvements, showing the effectiveness of our positional prompt-tuning method. 

\input{tabs/ablation}

\subsection{Few-shot Learning}
We also conduct experiments of few-shot learning on the ModelNet40~\cite{ModelNet15} dataset. We consider settings of n-way k-shot manner following previous works~\cite{PointMAE, recon23}, where n $\in$ 5, 10 and k $\in$ 10, 20. As shown in \cref{tab:few-shot}, it can be observed that our PPT can achieve performance enhancement in most cases compared with IDPT~\cite{IDPT} and DAPT~\cite{DAPT}, which also indicates the effectiveness of our method.

\subsection{3D Part Segmentation}
To evaluate the geometric understanding performance within objects, we conducted the part segmentation experiment on ShapeNetPart~\cite{ShapeNetPart16}. 
Specifically, we concatenate the cross-modal feature into the global feature and use the same segmentation head as Point-MAE for a fair comparison. 
From \cref{tab:partseg}, it can be observed that our method gets results on par with the \textit{from scratch} baseline on both Cls. mIoU and Inst. mIoU. 
Besides, PPT also outperforms the PEFT counterparts IDPT~\cite{IDPT} and DAPT~\cite{DAPT} by 0.06\% and 0.46\% Cls. mIoU, respectively. 
It shows that the cross-modal global knowledge from cross-modal pretraining can still play a certain role in part segmentation.

\subsection{Ablation Studys}
We conduct ablation experiments to demonstrate the effectiveness of each module in our method. Details and results are provided as follows.

\smallskip
\noindent\textbf{Ablation study on position embedding forms.}
In \cref{tab:4}, we replace the positional embedding layer in the PPT with several classic forms. As we can see in the table, the results show that such MLP structures as position embedding layers, as we mentioned before, perform better than others, which also proves that the point Transformer with MLP embedding layers serves as a multi-scale feature extractor. It's surprising that tuning with only the first layer receives MLP-embedded positional information can achieve relatively high performance, even better than all other forms. 

\smallskip
\noindent\textbf{Ablation on group number K.}
We conduct an ablation study on the number of patch token groups K, ranging from 1 to 3. The best choice is K=2, showing the effectiveness of extra prompt tokens. Too many prompt token groups can interfere with fine-tuning. However, K=3, though higher than K=1, still demonstrated the benefit of extra token prompts.

\input{figs/posnet}
\input{tabs/posnet}

%% file: tabs/segmentation.tex
\begin{table*}[h!]
\caption{\textbf{Part segmentation on ShapeNetPart dataset}. The number of training parameters (M), mIoU over all classes (Cls.) and the mIoU over all instances (Inst.) are reported.
}
\label{tab:partseg}
\centering
\setlength\tabcolsep{6pt}
\begin{tabular}{lcccc}
\toprule
Methods & Reference & \#TP (M)& Cls. mIoU (\%) & Inst. mIoU (\%) \\
\midrule
\multicolumn{5}{c}{\textit{Supervised Learning Only}} \\
\midrule
PointNet \cite{PointNet} & CVPR 17  &- & 80.39 & 83.7 \\
PointNet++  \cite{PointNet++}  & NeurIPS 17 &-  & 81.85 & 85.1 \\
DGCNN \cite{DGCNN} & TOG 19 & - & 82.33 & 85.2 \\
\midrule
\multicolumn{5}{c}{\textit{ Self-Supervised Representation Learning (Full fine-tuning)}} \\
\midrule
OcCo \cite{occo} & ICCV 21 & 27.09 & 83.42 & 85.1 \\
Point-BERT \cite{PointBERT} & CVPR 22 & 27.09 & 84.11 & 85.6 \\
Point-MAE \cite{PointMAE} & ECCV 22 & 27.06 & 84.19 & 86.1 \\ 
\act \cite{ACT23} & ICLR 23 &  27.06 & 84.66 & 86.1 \\
\recon \cite{recon23} & ICML 23 &  27.06 & 84.66 & 86.4 \\
\midrule
\multicolumn{5}{c}{\textit{with Parameter-Efficient Supervised Fine-tuning}} \\
\midrule
Point-MAE~\cite{PointMAE}~(baseline) &  ECCV 22 & 27.06 & 84.19 & 86.1 \\ 
+ IDPT~\cite{IDPT} & ICCV 23 & 5.69  & 83.79  & 85.7  \\
+ DAPT~\cite{DAPT} & CVPR 24 & 5.65  & 84.01 & 85.7 \\
\rowcolor{table_row}+ \ppt~(ours) & - &\textbf{5.62} & \textbf{84.07} & \textbf{85.7}\\
\midrule
\recon~\cite{recon23}~(baseline) & ICML 23 & 27.06 & 84.52 & 86.1 \\
+ IDPT~\cite{IDPT} &  ICCV 23 & 5.69 & 83.66  & \textbf{85.7} \\
+ DAPT~\cite{DAPT} & CVPR 24 & 5.65  &83.87 & 85.7\\
\rowcolor{table_row}+ \ppt~(ours) & ACMMM 25 &\textbf{5.62} & \textbf{84.23} & 85.6 \\
\bottomrule
\end{tabular}
\vspace{5pt}
\end{table*}

%% file: tabs/ablation.tex
\begin{table*}
\centering
\scriptsize
\begin{minipage}[h]{0.51\textwidth}
\centering
\caption{\textbf{Ablation of different position embedding forms.} First denotes that only the first layer gets positional information.}
\vspace{-2pt}
\label{tab:4}
\resizebox{0.9\linewidth}{!}{
\setlength\tabcolsep{5.5pt}
\begin{tabular}{lcc}
    \toprule
    Position Encoding Forms & Position & PB\_T50\_RS \\
    \midrule
    PPT + Cosine & All & 88.45 \\
    PPT + RoPE~\cite{su2024roformer} & All & 88.58 \\
    PPT + NeRF~\cite{nerf21} & All & 88.69 \\
    PPT + MLP & First & 88.69 \\
    \rowcolor{table_row} PPT + MLP & All & 89.52  \\
    \bottomrule
\end{tabular}
}
\end{minipage}
\hspace{25pt}
\begin{minipage}[h]{0.4\textwidth}
\makeatletter\def\@captype{table}\makeatother
\centering
\setlength\tabcolsep{8pt}
\caption{\textbf{Ablation on the number of patch token groups K.} When K=2, the algorithm achieves optimal performance and strikes a balance in terms of the number of training parameters.}
\label{tab:5}
\vspace{-2pt}
\resizebox{0.9\linewidth}{!}{
\begin{tabular}{ccc}
    \toprule
    Num K &\#TP (M) &PB\_T50\_RS \\
    \midrule
    1 & 0.9M &  87.40 \\
    \rowcolor{table_row} 2 & 1.1M &  89.52 \\
    3 & 1.4M & 88.06  \\
    \bottomrule
\end{tabular}
}
\end{minipage}
\end{table*}
\vspace{8pt}

%% file: figs/posnet.tex
\begin{figure}[t] 
    \centering  
    \includegraphics[width=\linewidth]{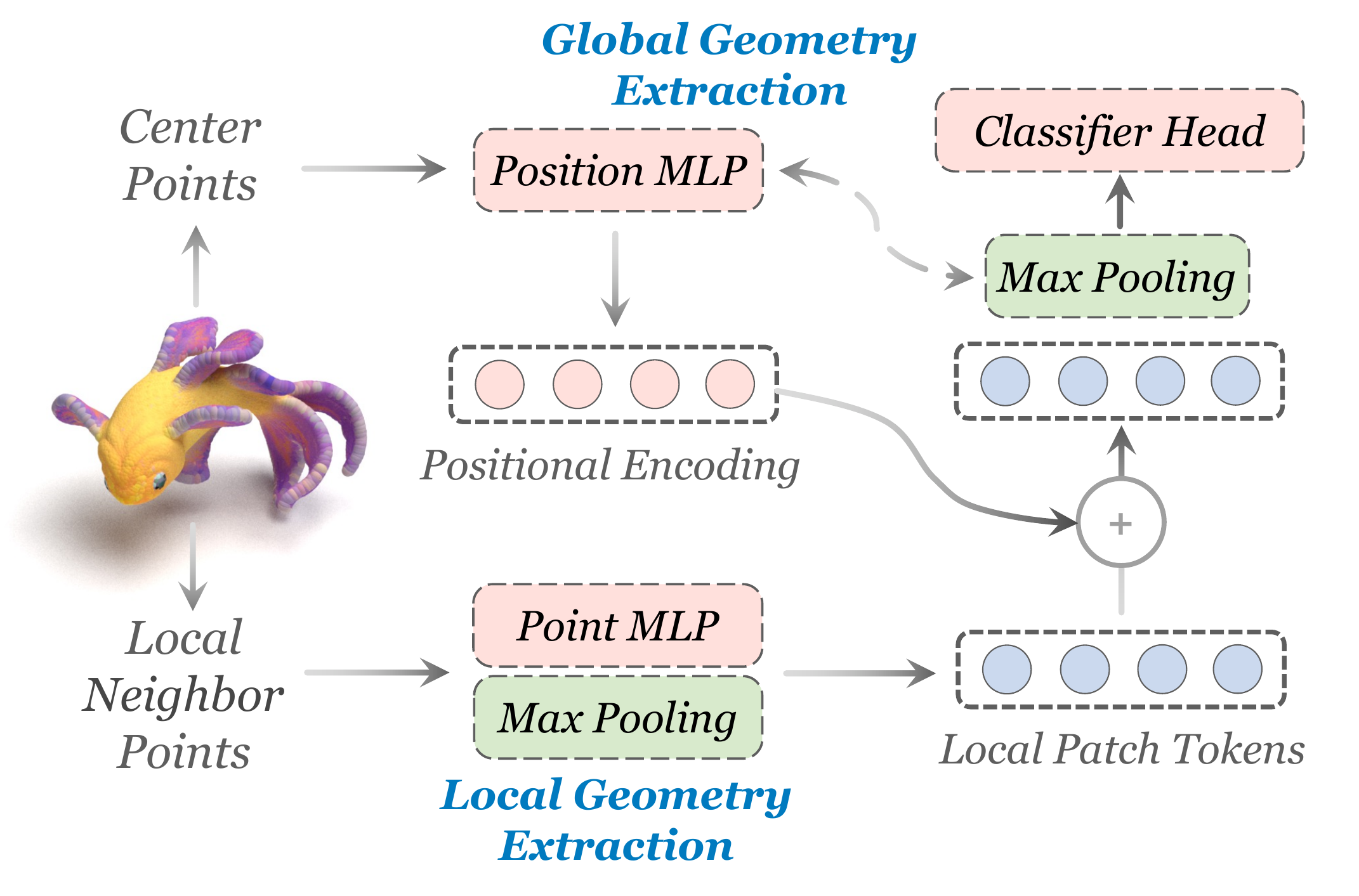}
    \vspace{-18pt}
    \caption{\textbf{Pipeline of PosNet}, a simple structure only has \textbf{0.8M} parameters. It only includes MLP and pooling layers, without any Transformer blocks.}
    \label{fig:posnet}
\end{figure}

%% file: tabs/posnet.tex
\begin{table}[t!]
\caption{\textbf{Classification preference of PosNet on ScanObjectNN.} We don't use any additional data for Pre-training.
}
\vspace{-6pt}
\label{tab:pos}
\centering
\setlength\tabcolsep{1.5pt}
\begin{tabular}{lcccc}
    \toprule
    Methods & Params. (M) & OBJ\_BG& OBJ\_ON & PB\_T50\_RS \\
    \midrule
    PointNet~ & 3.5 & 73.3 & 79.2 & 68.0 \\
    PointNet++ & 1.5 & 82.3 & 84.3 & 77.9\\
    MVTN & 11.2 & 92.6 & 92.3 & 82.8\\
    Point-BERT & 22.1  & 87.43 & 88.12 & 83.07 \\
    \rowcolor{table_row}PosNet (Ours) & \textbf{0.8} & \textbf{90.02} & 8\textbf{9.28} & \textbf{84.41} \\
    \bottomrule
\vspace{-0.5cm}
\end{tabular}
\end{table}

%% file: chapters/discussions.tex
\section{Discussions}
\subsection{Position is all your need in 3D Transformer?}
Positional encoding is particularly crucial in 3D Transformers. We have demonstrated that combining MLP-based positional encoding with local patch tokens effectively functions as both global and local point cloud feature extractors (e.g., PointNet). A bold hypothesis is whether, without the Transformer blocks, positional encoding tokens alone can achieve comparable or even superior performance.

\cref{fig:posnet} illustrates this very simple architecture, and its classification performance on ScanObjectNN is shown in \cref{tab:pos}. With only \textbf{0.8M} parameters, PosNet achieved remarkable performance, further validating our theoretical assumption that the global understanding provided by positional encoding is crucial in 3D Transformers.

\input{figs/attn_visual}
\subsection{Attention Visualization in Position Encodings}
We also report the attention score visualization of either position embedding or patch tokens as the input, respectively. Taking the 3D coordinates of patch centers as the input of position embedding, the attention scores of the position embedding input are more holistic, as shown in the top row of \cref{fig:attn_visual}. That means the position embedding features are concerned with more global information. Meanwhile, the patch token embedding, with all neighbor points around the cluster centers as input, gathers together their attention gathered together at some points, as the local counterpart of position embedding inputs. Concerning both global and local features, the point cloud Transformer architectures deal with complex tasks with a simple MLP as the position encoding module and even get better results than adding those well-designed position embedding methods together, as reported in \cref{tab:4}.

%% file: figs/attn_visual.tex
\begin{figure}[t] 
    \centering  
    \includegraphics[width=\linewidth]{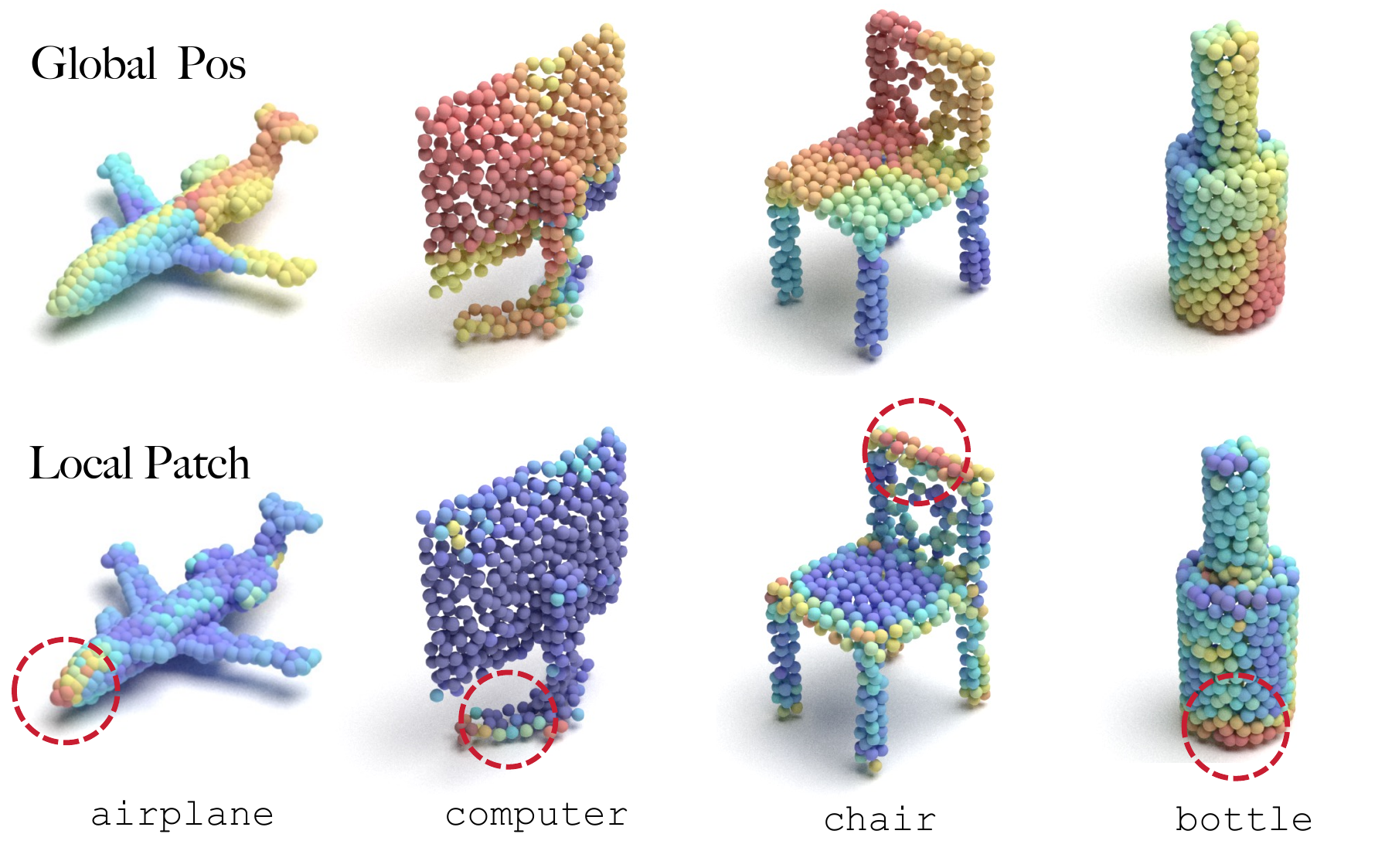}
    \vspace{-17pt}
    \caption{The visualization of the attention of position embedding and patch tokens. The top row represents the visualization results of position embedding inputs, and the bottom row shows the results of patch token inputs.}
    \label{fig:attn_visual}
\end{figure}

%% file: chapters/future_work.tex
\section{Limitations \& Future Work}
Although we want to underscore the significance of positional encoding within 3D Transformer and have experimentally validated the use of positional encoding as a form of prompt tuning in classic 3D self-supervised pre-training works, such as Point-MAE~\cite{PointMAE} and ReCon~\cite{recon23}, we have yet to conduct experimental validation on larger datasets, such as Objaverse~\cite{objaverse23}. Our primary future work will be to experiment with positional prompt tuning on a wider range of 3D foundational models, such as OpenShape~\cite{OpenShape23} and ReCon++~\cite{shapellm24}, and to expand its application to a broader spectrum of tasks, including zero-shot learning~\cite{ULIP22,OpenShape23,recon23,shapellm24,xie2024fusing,lin2025peva}, 3D AIGC~\cite{pointe22,vpp23,ma2024large}, 3D large language models~\cite{shapellm24,dreamllm24,tang2025exploring,xiong20253ur}, and robotics~\cite{sofar25,dreamvla25,omnispatial25,dexvlg25}, among others~\cite{DGMS22,dreambench++24,PointDistiller22,yu2024seqafford}. There are also plenty of new methods related to 3D representation learning or parameter-efficient fine-tuning~\cite{wang2024rethinking,kang2024point,liu2025pointcg,sun2025learnable,sun2024exploiting}, which we will conduct experiments on.

%% file: chapters/conclusion.tex
\section{Conclusions}
In this paper, we revisit the role of position encoding in 3D Transformer models, highlighting its simplicity and importance as a high-dimensional element. By combining position encoding with the patch encoder, we capture multi-scale information. This, alongside the sequential Transformer, creates a comprehensive framework that integrates both local features from patches and global features from center points through position encoding. With just a few parameters, the position embedding module aligns well with PEFT tasks, making it ideal for fine-tuning. Extensive experiments validate that PPT is both effective and efficient.
\section*{Acknowledgments}
This work was sponsored by the National Natural Science Foundation of China No. 62472349, the Fundamental Research Funds for the Central Universities No. xxj032023020, and CAAI-CANN Open Fund, developed on OpenI Community.

%% file: chapters/appendix.tex
\newpage
\appendix
\onecolumn
\section{Implement Details}
\subsection{Training recipes}
\input{tabs/details}
To ensure a fair comparison, we employed identical experimental settings to the default fine-tuning method for each baseline. This entails freezing the weights of the pre-trained point cloud backbone and solely updating the newly inserted parameters during training. All experiments are conducted on a single GeForce RTX 3090. In \cref{tab:hyper_params}, we show all our training details.

\section{Additional Ablation Study}

\subsection{Ablation on different settings of MLP elements.}
Our \ppt~contains a number of different MLP components. For the significance of these components, we conducted comprehensive ablation experiments to prove the effectiveness of these parts. As shown in \cref{tab:mlp_ablation}, we carry out experiments on different structures with the same training settings. Specifically, we test the performance on 3 subsets of Scan\_Object\_NN dataset (PB\_T50\_RS, OBJ\_BG, OBJ\_ONLY). Both Point-MAE and \recon~are considered as baselines. In detail, we choose three different structural settings for ablation: removing the adapter MLP, freezing the weight of the Point MLP during training, and freezing the weight of the Position MLP during training. And our results are listed in the last column. It is worth noting that all other model Settings are consistent with those of \ppt when these changes are made. 
    
We can tell from the results that all these different changes in model parts result in significant performance reduction, which laterally reflects their effectiveness. Respectively, the addition of the adapter MLP brings the most improvement (3.15\% avg. of Point-MAE and 2.65\% avg. of \recon). During fine-tuning, either freezing point MLP (0.77\% avg. of Point-MAE and 0.83\% avg. of \recon) or position MLP (0.58\% avg. of Point-MAE and 0.90\% avg. of \recon) renders a performance reduction.

\input{tabs/add_ablation}

\section{Additional Visualization}
\subsection{t-SNE Visualization}
\input{figs/tsne}
In \cref{fig:tsne}, we report the \textbf{t-SNE} visualization results of Point-MAE\cite{PointM2AE22}, DAPT\cite{DAPT}, and our \ppt. Here, we report the point cloud feature manifold visualization results on the ScanObjectNN PB\_T50\_RS dataset. It can be observed that our method, with fewer or on par number of parameters, gets more compact clusters and more separate cluster centers, which means our method gets better results than the parameter-efficient full fine-tuning counterparts.

%% file: tabs/details.tex
\begin{table*}[h!]
\caption{\textbf{Training recipes for downstream fine-tuning tasks}.}
\vspace{-10pt}
\label{tab:hyper_params}
\vskip 0.10in
\centering
\begin{tabular}{lcccc} & \multicolumn{2}{c}{\texttt{Classification}} & \texttt{Few-Shot} & \texttt{Segmentation}\\
\toprule[0.95pt]
Config & ScanObjectNN & ModelNet & ModelNet & ShapeNetPart\\
\midrule[0.6pt]
optimizer & AdamW & AdamW & AdamW & AdamW\\
learning rate& 2e-5 & 1e-5 & 5e-4  & 2e-4 \\
weight decay& 5e-2 & 5e-2 & 5e-2  & 5e-2 \\
learning rate scheduler & cosine & cosine & cosine & cosine \\
training epochs & 300 & 300 & 150 & 300\\
warmup epochs & 10 & 10 & 10 & 10\\
batch size & 32 & 32 & 32 & 16\\
drop path rate & 0.3 & 0.3 & 0.2 & 0.1 \\
\midrule[0.6pt]
number of points & 2048 & 1024/8192 & 1024 & 2048 \\
number of point patches & 128 & 64/512 & 64 & 128 \\
point patch size & 32 & 32 & 32 & 32 \\
\midrule[0.6pt]
augmentation & Rotation & Scale \& Trans & Scale \& Trans & - \\
\midrule[0.6pt]
GPU device & RTX 3090 & RTX 3090 & RTX 3090 & RTX 3090 \\
\bottomrule[0.95pt]
\end{tabular}
\end{table*}

%% file: tabs/add_ablation.tex
\begin{table*}
\setlength\tabcolsep{7pt}
\caption{\textbf{Ablation study of different MLP element settings.} The optimal results are shown in bold. Here, no adapter MLP means removing adapter MLPs while training, \textit{freezing} point or pos MLP means freezing the weight of point MLP or the position MLP.}
\vspace{-5pt}
\label{tab:mlp_ablation}
\begin{tabular}{l|l|cccc}
    \toprule
    Method & SCAN\_OBJECT\_NN & no adapter MLP & \textit{freezing} point MLP &\textit{freezing} pos MLP  &PPT\\
    \midrule
    PointMAE~\cite{PointMAE} & PB\_T50\_RS & 85.39 & 87.89 & 87.93 & \textbf{89.00} \\
    PointMAE~\cite{PointMAE} & OBJ\_BG & 89.85 & 92.94 & 93.12 & \textbf{93.63} \\
    PointMAE~\cite{PointMAE} & OBJ\_ONLY & 90.53 & 92.08 & 92.43 & \textbf{92.60} \\
    \midrule
    \recon~\cite{recon23} & PB\_T50\_RS & 86.22 & 89.10 & 89.04 & \textbf{89.52} \\
    \recon~\cite{recon23} & OBJ\_BG & 92.08 & 93.46 & 94.49 & \textbf{95.01} \\
    \recon~\cite{recon23} & OBJ\_ONLY & 91.57 & 92.77 & 91.57 & \textbf{93.28} \\
    \bottomrule
\end{tabular}
\end{table*}

%% file: figs/tsne.tex
\begin{figure*}[h]
    \centering  
    \includegraphics[width=\linewidth]{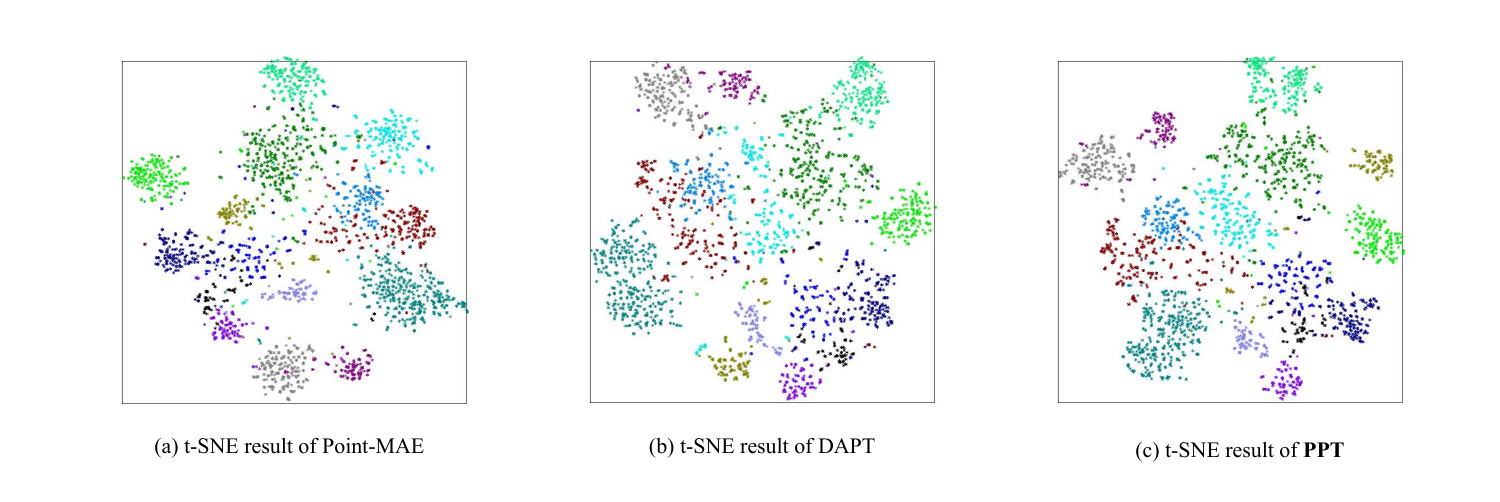}
    \vspace{-25pt}
    \caption{The t-SNE visualization results for Point-MAE, DAPT, and our \ppt. The Point-MAE result is obtained with a full fine-tuning paradigm, while others are all partially fine-tuned results.}
    \label{fig:tsne}
\end{figure*}